# Localized Traffic Sign Detection with Multi-scale Deconvolution Networks

Songwen Pei[1,2], Fuwu Tang[1], Yanfei Ji[1], Jing Fan[1], Zhong Ning[2]

[1] Department of Computer Science and Engineering
University of Shanghai for Science and Technology
Shanghai, China
swpei@usst.edu.cn

[2] School of Management
Fudan University
Shanghai, China
swpei@fudan.edu.cn

*Abstract*—Autonomous driving is becoming a future practical lifestyle greatly driven by deep learning. Specifically, an effective traffic sign detection by deep learning plays a critical role for it. However, different countries have different sets of traffic signs, making localized traffic sign recognition model training a tedious and daunting task. To address the issues of taking amount of time to compute complicate algorithm and low ratio of detecting blurred and sub-pixel images of localized traffic signs, we propose Multi-Scale Deconvolution Networks (MDN), which flexibly combines multi-scale convolutional neural network with deconvolution sub-network, leading to efficient and reliable localized traffic sign recognition model training. It is demonstrated that the proposed MDN is effective compared with classical algorithms on the benchmarks of the localized traffic sign, such as Chinese Traffic Sign Dataset (CTSD), and the German Traffic Sign Benchmarks (GTSRB).

## I. Introduction

Autonomous driving has been attracting the attentions from the academia as well as the industry. Vehicles equipped with sensors and cameras navigate in China, Japan, United States, Korea, Europe, almost all around the world [24, 25]. However, without sophisticated artificial intelligence models and high definition maps to analyze information and the capacity to learn from changing circumstances, autonomous vehicles would be difficult to operate safely [1]. The comprehensive and complex autonomous driving system involves broad technologies including sensing, perception, localization, decision-making, real-time operating system, heterogeneous computing, graphic/video processing, as well as cloud computing, etc.

Object recognition is one critical step in the perception supported by the intelligent object collector. Object recognition and tracking can be achieved with Deep Learning, which, in recent years, has been a rapid development. It implements accurate object detection and tracking using camera inputs. A convolutional neural network (CNN) is a type of deep neural network that is widely used in object recognition tasks [2,3].

One practical challenges faced by autonomous driving industry is localized traffic sign recognition. The reason is that each country has its own set of traffic sign, thus re-training traffic sign recognition models for each country is a tedious and daunting task. In this paper, to address this challenge, we propose and implement Multi-Scale Deconvolution Networks (MDN), which flexibly combines multi-scale convolutional neural network with deconvolution sub-network, thus leading to efficient and reliable localized traffic sign recognition model training.

Specifically, due to the difficulties of detecting traffic sign, such as the variety of traffic sign, illumination, blurred color images, traffic regulations and so on, it is a big challenge to recognize a traffic sign in real time and make a right decision for driver throughout all the countries and regions. However, since traffic sign are broadly made with specific colors, size, shapes, and there are high similarity in shapes, sizes, and colors, it could be detected by using the physical attributes of colors, shapes information.

As a result, color based detection, shape based detection and both of color and shape detection algorithms are broadly utilized in the traditional recognition applications based on the methods of computer graphics. However, as the rapid development of deep learning based algorithms on graphic recognition, it would be a potential alternative to be implemented in autonomous driving system.

Generally, to recognize traffic sign images, the autonomous driving system contain at least two steps: Classification and Detection [4]. In consequence, in the proposed MDN framework, the recognition system of traffic sign consists of three stages. First, we train the classification stage with classic Convolutional Neural Network (CNN) and advanced residual network on the German Traffic Sign Benchmarks (GTSRB)[5]. Then, in order to take an example on localized traffic sign recognition, we refine and make the decision of recognizing on image dataset of traffic sign Chinese Traffic Sign Dataset

(CTSD) [6] by adding redundant deconvolution networks. Finally, we detect the multi-scale images derived from the upper levels, and integrate the multi-way detectors.

The main contribution of this paper are as follows: (1) a unified framework MDN combining classical convolutional neural network with deconvolution network can effectively train multi-scale images of traffic sign; (2) two different traffic sign datasets collecting German and Chinese sign images are trained and compared; (3) a relatively small amount of traffic sign images from long distance taken are involved to be detected; (4) especially, some of traffic sign covered by environmental objects, such as trees, leaves, and others are also been labeled and detected.

This paper is organized as follows: Section II briefly introduce the related work on traffic sign detection and deep learning frameworks, Section III expatiates the proposed multi-scale deconvolution network, Section IV shows the experiments details and compares the experimental results, Section V makes the conclusions.

## II. RELATED WORK

### A. Traditional Methods

In the traditional method for recognition traffic signs mainly base on the feature extraction algorithms. Maldonado-Bascon et al. transformed the image into the HSI color space and calculated histogram of Hue and Saturation components. Then, the histogram was classified by using a multiclass SVM[14]. Zaklouta et al. used the Histogram of Oriented Gradient (HOG) to extract image feature, and used a SVM for image classification[15]. In order to reduce the memory and enhance the performance, they utilized a Fisher's criterion and random forests for features extraction. Wang et al. proposed a new algorithm to detect circle for matching traffic signs[16]. Experimental results show that it is much more efficient to match the traffic signs and it can reduce the computation by less memory access. Yang et al. proposed color probability model to transform traffic sign images to probability maps, and an integral channel features detector was employed to remove false positives of the proposal[17]. Compared with the state-of-the-art methods the algorithm can improves the computational efficiency. Wang et al. propose a robust ellipse-detection method based on sorted merging[18].

### B. Deep learning methods

Deep learning has been popular to detecting images. Jin et al. proposed a new neural networks architecture and utilized the rectified linear unit (ReLU) activations to recognize the traffic signs[19]. To train neural networks model, they proposed a stochastic gradient descent (HLSGD) method. They gained a recognition rate of 99.65% on the German Traffic Sign Recognition Benchmark(GTSRB). Cire et al. proposed a Multi-column Deep Neural Networks for image classification [1]. The neural networks architecture combining various DNNs trained on different preprocessed data into a Multi-Column DNN (MCDNN) which achieved a better-than-human recognition rate of 99.46% on the GTSRB. Zhang et al. propose a Chinese traffic sign detection end-to-end convolutional network inspired by YOLOv2[20]. In order to reduce the top layers computational complexity, they took the multiple 1*1 convolutional layers and divided the input images into dense grids to detect traffic signs. Therefore, they got better performance and detected small traffic signs, and the detection time was only 0.017 second on average for each image. W Liu et al. presented a multi-scale deconvolution networks method to detect objects [21]. It combined predictions object from multiple feature maps and use default boxes from multiple layers. Compared with the YOLO[22] or Faster R-CNN[23], or even with a smaller input image size, SSD achieved higher performance. Lin T Y et al. proposed a feature pyramid networks(FPN) to merge multiple feature layers and detect objects at different scales[8]. The FPN networks achieves state-of-the-art from the COCO 2016 challenge winners. Hengliang Luo et al. designed a new data-driven system to recognize all categories of traffic signs[4]. The system can recognize all categories of traffic signs, which contains three stages. The first stage is to extract traffic sign from the regions of interest (ROIs) based on symbol and text, then refine ROIs, lastly it use a neural network to implement classification. Lim K et al. presented a real-time traffic sign detection method based on General Purpose Graphics Processing Unit (GPGPU)[26]. Sermanet et al. utilized a sliding window for convolutional networks classification and object detection[27]. It was the first time to use convolutional networks to predict object boundaries, and they gained the winner of the ImageNet Large Scale Visual Recognition Challenge 2013.

The traditional detection methods have advantages in accuracy but they are complicate to be implemented in time-demanding tasks. The convolutional neural network that is able to analyze high-resolution images in real time. Therefore, we propose a multi-scale deep neural networks framework by fusing convolutional neural network with residual network and modifying feature pyramid networks.

## III. MULTI-SCALE DECONVOLUTION NETWORK

The proposed multi-scale deconvolution network (MDN) is combined with three stages, as shown in the Fig.1. MDN mainly consists of Convolutional Residual Network (CRN), modified Feature Pyramid Network (MFPN), and the multi-scale classifier and detector.

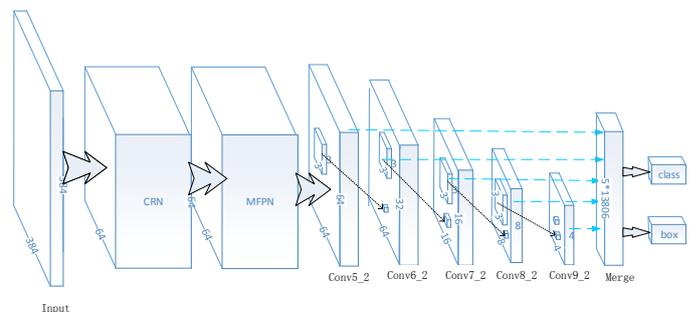

Fig.1 Framework of Multi-scale deconvolution network(MDN)

The sequences of traffic sign images or frames are input into the MDN at the size of 384 by 384. Then, it will be passed into the first stage to extract features of traffic sign. In this stage, we advanced the classic convolutional neural network with two levels of residual networks(CRN). Thus, we can classify the abundant of traffic sign images and train the parameters of CRN. After then, we modify the Feature Pyramid Network (FPN) to detect objects. It derives from the deconvolution network and refines small size of feature images on the two opposite ways.

This pyramid hierarchy can extract features of different spatial resolutions, which helps with improving the accuracy and efficiency of recognizing the blurred small size images or low-resolution features in real time. Then, we define 5 different scales for traffic features and implement them with 5 filters Conv5_2, Conv6_2, Conv7_2, Conv8_2, and Conv9_2. Finally, the softmax layer will generate the results of classes, and corresponding locations. The label of class includes the quantity of classes, and the label of box predict the location information of each corresponding class.

While training the image features in the framework of MDN, we utilize 5 default boxes similar to the concept from SSD Network [7]. Each feature image can be divided into several units with size of M*N. For example, the size of feature image is 48*48 in Conv5_2. Each default box can predict some classes.

Similarity, we define the overall loss function as the weighted loss function in [7]. The loss function includes two parts: confidence loss(conf) and localization loss(loc).

$$L(x,c,l,g) = \frac{1}{N}(L_{conf}(x,c) + \alpha L_{loc}(x,l,g)) \quad (1)$$

where N is the number of default boxes. In the MDN framework, the N is equal to 5. The parameter $\alpha$ is the weighted value to balance the effects of confidence loss and localization loss.

As defined in [7], the confidence loss function and localization loss function are also define as follows:

$$L_{conf}(x,c) = -\sum_{i \in Pos}^{n} x_{ij}^{p} \log(\hat{c}_i^p) - \sum_{i \in Neg}^{n} \log(\hat{c}_i^0) \quad (2)$$

where $c_i^p = \frac{\exp(c_i^p)}{\sum_p \exp(c_i^p)}$.

For example, $x_{ij}^p = \{1,0\}$ means that the i[th] default box matches ground truth box $P$ in the j[th] class.

The localization loss function is accelerating sum of Smooth loss between the predicted box and the ground truth box(g).

$$L_{loc}(x,l,g) = \sum_{i \in Pos}^{N} \sum_{m \in \{xc,cy,w,h\}} x_{ij}^k smooth_{L1}(l_i^m - \hat{g}_j^m) \quad (3)$$

where $\hat{g}_j^{cx} = (g_j^{cx} - d_i^{cx})/d_i^w$, $\hat{g}_j^{cy} = (g_j^{cy} - d_i^{cy})/d_i^h$,

$\hat{g}_j^w = \log(\frac{g_j^w}{d_i^w})$, $\hat{g}_j^h = \log(\frac{g_j^h}{d_i^h})$.

Currently, we implement the same definition of overall loss function, confidence loss function and localization loss function as in [7]. In the next steps, we will extend it by adding new parameters, such as the ratio of covered ground truth *cg*, the depth of traffic sign *dg*, and the shape class of image *sc*. These parameters would accelerate the speed of converging classification and improve the accuracy of recognizing on different levels.

The most effective capability of MDN is to recognize the sub-pixel traffic sign and to achieve the comparable performance of generative adversarial networks for blurred and small size images from long distance depth image.

A. *Convolutional Residual Network*

The Convolutional Residual Network (CRN) is an enhanced state-of-art Convolutional Neural Network (CNN), as shown in the Fig.2. The CRN consists of two-level residual network to enhance CNN. In the model of CRN, the size of each convolutional kernel is 3 by 3. The output of maxpool layer from the classic CNN will be input into the first level residual network in which conv2 and conv2_2 are fused as the next layer of maxpool. At the second level residual are done the similar schema by decreasing the size of feature image from 16 by 16 to smaller size of 4 by 4.

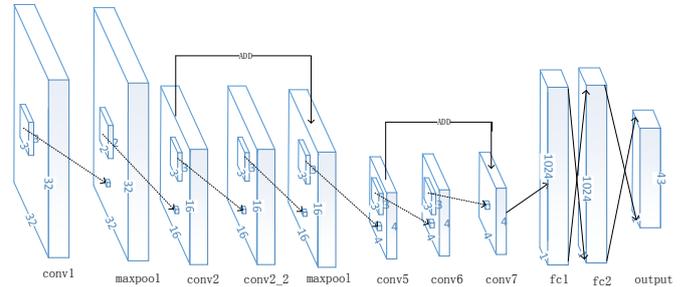

Fig.2 Convolutional Residual Network(CRN)

Then, the output of two full connection layers will go into the softmax layer. In order to accelerate the speed of executing convolutional kernel, we normalize the feature image in each layer of convolution by the function of Batch Nornalization.

Due to the two extra residual networks on different size of feature images, we are able to build short cuts to fasten convergence. Therefore, the CRN will be avoid of abundant computation for more convolutional layers.

B. *Modified Feature Pyramid Network*

Feature pyramid network [8] was proposed to detect objects at different scales by a top-down architecture which connected among high-level semantic feature maps at all scales. It gains high performance of producing a single high-level feature map to predict a high-resolution one from blurred and smaller size feature map. Therefore, it can help to generate small objects with poor representations.

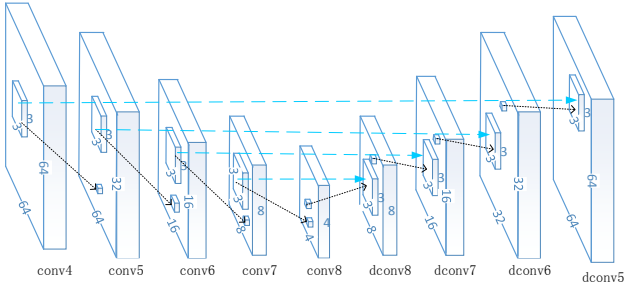

Fig.3 Modified Feature Pyramid Network

We learn from the pyramid architecture to propose Modified Feature Pyramid Network (MFPN) to infer new and small object from blurred traffic sign or partially covered images.

As shown in the Fig.3, we set the stride between *conv4* and *conv5* as 2, and the size of convolutional kernel is also 3 by 3. The size of feature image from *conv8* is 4 by 4. The conv8 is generated by fusing convolutional kernel of *conv7* and the up-level *conv6*. By operating convolutions 5 times later, we will start the corresponding deconvolution networks with 5 layers. The main benefit of deconvolution structure can achieve the similar capability of a smaller perceptual generative adversarial network. Therefore, it enhance the accuracy of recognition small objects from the image of traffic sign, especially those depth images taken from long distance.

### C. Multi-scale Classifier

As the multi-channels of feature maps are generated by CRN and MFPN in pipeline, a softmax layer performs the function of multi-scale classifier by integrating the information of locations and classes.

## IV. EXPERIMENTATIONS AND EVALUATIONS

We implement the experiments with two steps. Firstly, we train the CRN model and generate the class libraries of traffic sign on the German Traffic Sign Benchmarks (GTSRB). Then, we recognize the small objects by using the MFPN on the Chinese Traffic Sign Dataset (CTSD). The classes of traffic sign between German and China are almost the same, so we expect to generate more classes from the larger dataset of GTSRB than the smaller one.

### A. Parameters of CRN

According to the structure of CRN, we set the corresponding parameters in terms of the Table I which defines the size of input map, the size of output map and the size of convolutional kernel. The parameters of each layer for the model of CRN according to the Table I.

TABLE I THE KEY PARAMETES OF EACH LAYER IN CRN

| Layers | Input size | Output size | Kernel size |
|---|---|---|---|
| Conv1 | 32*32*3 | 32*32*64 | 3*3 |
| maxpool | 32*32*64 | 16*16*64 | |
| Conv2_1 | 16*16*64 | 16*16*128 | 3*3 |
| Conv2_2 | 16*16*128 | 16*16*128 | 3*3 |
| maxpool | 16*16*64 | 8*8*64 | |
| Conv3_1 | 8*8*256 | 8*8*256 | 3*3 |
| Conv3_2 | 8*8*256 | 8*8*256 | 3*3 |
| Conv3_3 | 8*8*256 | 8*8*256 | 3*3 |
| Fc_layer1 | 8*8*256 | 1024*1024 | |
| Fc_layer2 | 1024*1024 | 1024*1024 | |
| Fc_layer2 | 1024*1024 | 1024*43 | |

### B. Parameters of deconvolutions

We also set the parameters of each layer for the model of MFPN according to the Table II.

TABLE II THE KEY PARAMETES OF EACH LAYER IN MFPN

| Layers | Input size | Output size | Kernel size | stride |
|---|---|---|---|---|
| Conv4 | 64*64*256 | 32*32*256 | 3*3 | 2 |
| Conv5 | 32*32*256 | 16*16*256 | 3*3 | 2 |
| Conv6 | 16*16*512 | 8*8*512 | 3*3 | 2 |
| Conv7 | 8*8*512 | 4*4*1024 | 3*3 | 2 |
| Conv8 | 4*4*1024 | 2*2*1024 | 3*3 | 2 |
| Dconv8 | 4*4*512 | 8*8*256 | 3*3 | 2 |
| Dconv7 | 8*8*256 | 16*16*256 | 3*3 | 2 |
| Dconv6 | 16*16*256 | 32*32*256 | 3*3 | 2 |
| Dconv5 | 32*32*256 | 64*64*256 | 3*3 | 2 |

### C. Parameters of detection

We also set the parameters of each layer for the Multi-scale detection according to the Table III.

TABLE III THE KEY PARAMETES OF EACH LAYER IN MFPN

| Layers | Input size | Output size | Kernel size | stride |
|---|---|---|---|---|
| Conv5_2 | 64*64*256 | 32*32*256 | 3*3 | 2 |
| Conv6_2 | 32*32*256 | 16*16*256 | 3*3 | 2 |
| Conv7_2 | 16*16*256 | 8*8*512 | 3*3 | 2 |
| Conv8_2 | 8*8*512 | 4*4*512 | 3*3 | 2 |
| Conv9_2 | 4*4*512 | 2*2*512 | 3*3 | 2 |

### D. Training Classes

GTSRB includes 43 classes traffic sign, its total images are 39209, and the sample images under test are 12630. During the experiments, we expand the images in each class by data augmentation techniques into 500 samples, if the total sample of a class is less than 500; otherwise, we will expand the sample images to 1000 if the amount of sample data is less than 1000. Besides, we uniform the size of all sample images to 32 by 32.

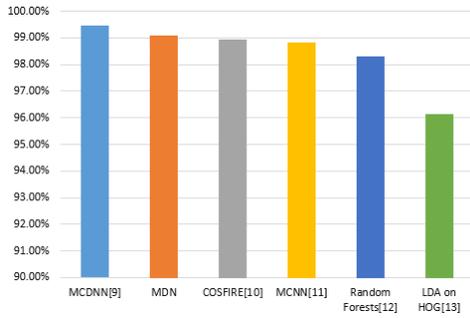

Fig. 4  Accuracy of classifing traffic sign

We got the accuracy of classifying traffic sign images up to 99.1% by executing 200 iterations of CRN. As shown in the Fig.4, we can see that the CRN network structure in MDN has achieved the second highest rank of recognition accuracy for image classification. However, it does not achieve the highest accuracy compared with the MCDNN, which may be less iterations and the convolution layer to be executed while we train the limit datasets in the limit time. Compared with other algorithms, the CRN structure shows it advantages on recognition accuracy.

*E. Detection*

In this paper, we use the CTSD dataset[6] to detect Chinese traffic signs. The dataset contains 1100 images, 700 ones for training and 400 ones for testing. The dataset can be divided into three classes: Prohibitory signs, Mandatory signs, and Danger signs. In the three classes, there are 48 categories of traffic signs. The Prohibitory signs consist of red color and circular shape. The Mandatory signs consist of blue color and circular shape, and the Danger signs consist of yellow color and triangular shape. The image sizes are between 1024*768 and 1270*800. The sizes of the traffic signs for training vary from 20*20 to 380*378, and the sizes of the traffic signs for testing vary from 26*26 to 573*557. The training set contains 236 Danger signs, 540 Prohibitory signs and 266 Mandatory signs. The testing set contains 139 Mandatory signs, 129 Danger signs and 264 Prohibitory signs.

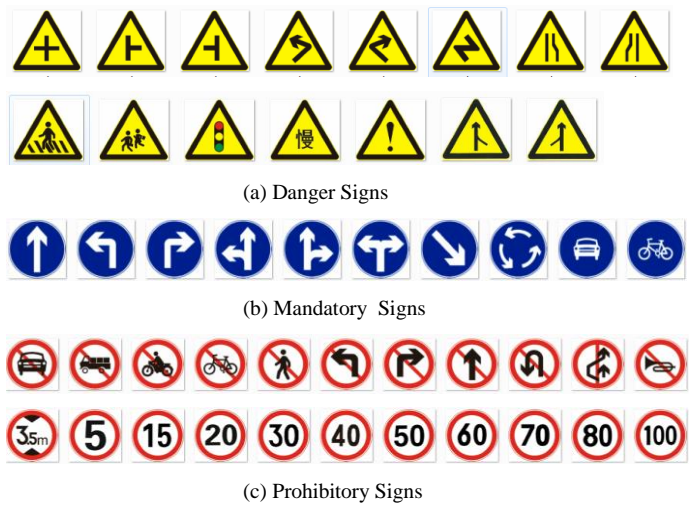

(a) Danger Signs

(b) Mandatory Signs

(c) Prohibitory Signs

Fig.5 Three classes of traffic signs

The three classes traffic signs are shown in the Fig.5. In the experimental results, we compared with SSD300 and set the initial threshold *t=0.5* to ensure the accuracy of the detection, where *t* indicates the confidence scores of the classes. When the detected category probability below the threshold *t*, we filtered those candidate areas of the traffic sign. The accuracy of detecting traffic sign are shown in the Fig.6.

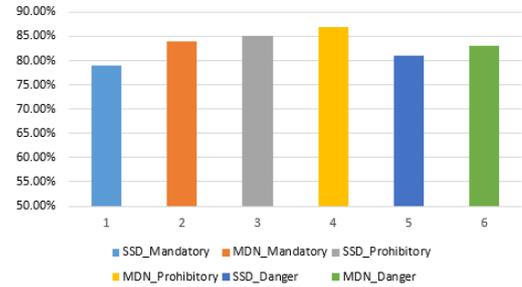

Fig.6 Accuracy of detecting traffic sign

As shown in the Fig.6, the MDN network performed better than SSD300. The accuracy of Mandatory sign performed by MDN is 6% higher than that of SSD300, and we also gain respective 2% higher accuracy on the Prohibitory sign and Danger sign than that of SSD300. Nevertheless, the average detecting time of MDN is a little bit slower than that of SSD300.

*F. Experimental evaluation*

The experiments on traffic sign detection and recognition are evaluated on CTSD. We evaluate different scenes. For example, in a given object image, it contains multiple traffic signs, the blurred traffic signs in fog, and complex scenes detection. The test results show more robust in real-world environments. Unfortunately, if some traffic signs covered by leaves or small object, they would not be detected in real time until now. The three scenes of experimental results are as shown in the Fig.7.

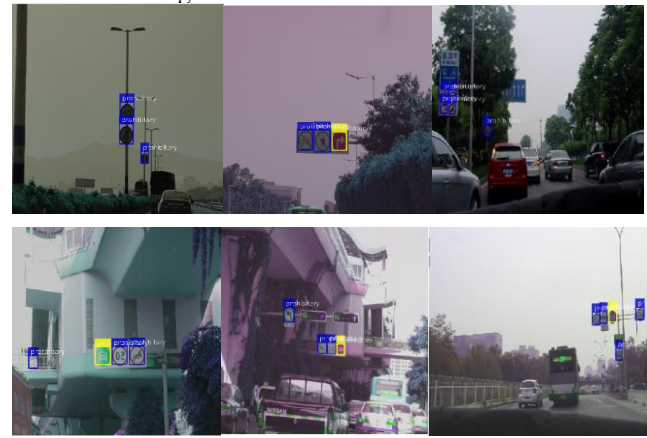

(a)  Multiple traffic signs in the object image

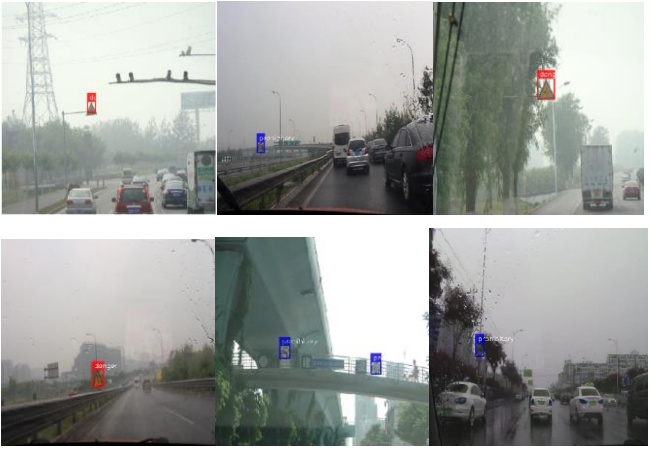

(b) multiple traffic signs in fog

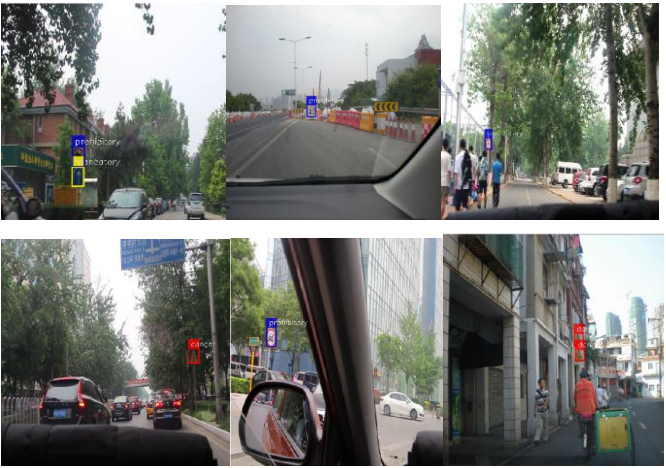

(c) complex scenes detection

Fig.7 Traffic sign detection in three scenes

## V. CONCLUSIONS

In this paper, we proposed a MDN network for real-time detection localized traffic sign based on CRN network and modified FPN for traffic sign classification. Especially, we use the modified FPN architecture and Multi-scale detection to achieve higher accuracy and accelerate the execution time on the specific localized traffic sign benchmark of Chinese Traffic Sign Dataset. The results confirm that the proposed network is an efficient and reliable way for localized traffic sign recognition. Especially on classification experiments of traffic sign images, the performance of CRN model was higher than most of other algorithms and the MDN model achieve better performance than SSD300. As a next step, we plan to enhance the Generative Adversarial Net to detect smaller size and blurred traffic signs taken from long distance.

ACKNOWLEDGMENT

We would like to thank the anonymous reviewers for their invaluable comments. This work was partially funded by the Shanghai Pujiang Program under Grant 16PJ1407600, the China Post-Doctoral Science Foundation under Grant 2017M610230, and the National Natural Science Foundation of China under Grant 61332009, 61775139. Any opinions, findings and conclusions expressed in this paper are those of the authors and do not necessarily reflect the views of the sponsors.